\documentclass[10pt,twocolumn,letterpaper]{article}
\usepackage{iccv}
\usepackage{times}
\usepackage{epsfig}
\usepackage{graphicx}
\usepackage{amsmath}
\usepackage{amssymb}
\usepackage{color}
\newcommand*\rot{\rotatebox{90}}
\usepackage{array}
\usepackage{booktabs}
\usepackage{pifont}
\usepackage{lscape}
\usepackage[table,xcdraw]{xcolor}
\usepackage{subfig}
\usepackage{xspace}
\usepackage{comment}
\usepackage{float}
\usepackage{epsfig}
\usepackage[table,xcdraw]{xcolor}
\usepackage{dblfloatfix}
\usepackage{bm}
\usepackage{iccv}
\usepackage{times}
\usepackage{epsfig}
\usepackage{graphicx}
\usepackage{amsmath}
\usepackage{amssymb}

\usepackage[breaklinks=true,bookmarks=false]{hyperref}

\iccvfinalcopy % *** Uncomment this line for the final submission

\def\iccvPaperID{****} % *** Enter the ICCV Paper ID here

\begin{document}
%\pagenumbering{gobble}
%%%%%%%%% TITLE
\title{Generating Visual Representations for Zero-Shot Classification}

\author{Maxime Bucher, St\'ephane Herbin\\
ONERA - The French Aerospace Lab\\
Palaiseau, France\\
{\tt\small maxime.bucher@onera.fr,}\\ {\tt \small stephane.herbin@onera.fr}
\and
Fr\'ed\'eric Jurie\\
Normandie Univ, UNICAEN,  ENSICAEN, CNRS\\
Caen, France\\
{\tt\small frederic.jurie@unicaen.fr}
}

\maketitle

\maketitle
\begin{abstract}

This paper addresses the task of learning an image classifier when some categories are defined by semantic descriptions only (\eg visual attributes) while the others are defined by exemplar images as well. This task is often referred to as the Zero-Shot classification task (ZSC). Most of the previous methods rely on learning a common embedding space allowing to compare visual features of unknown categories with semantic descriptions. This paper argues that these approaches are limited as i) efficient discriminative classifiers can't be used ii) classification tasks with seen and unseen categories (Generalized Zero-Shot Classification or GZSC) can't be addressed efficiently. In contrast, this paper suggests to address ZSC and GZSC by i) learning a conditional generator using seen classes ii) generate artificial training examples for the categories without exemplars. ZSC is then turned into a standard supervised learning problem. Experiments with 4 generative models and 5 datasets experimentally validate the approach, giving state-of-the-art results on both ZSC and GZSC.

\end{abstract}
\section{Introduction and related works}
\label{sec:related_work}
\begin{figure}[tb]
\centering
\includegraphics[width=8cm]{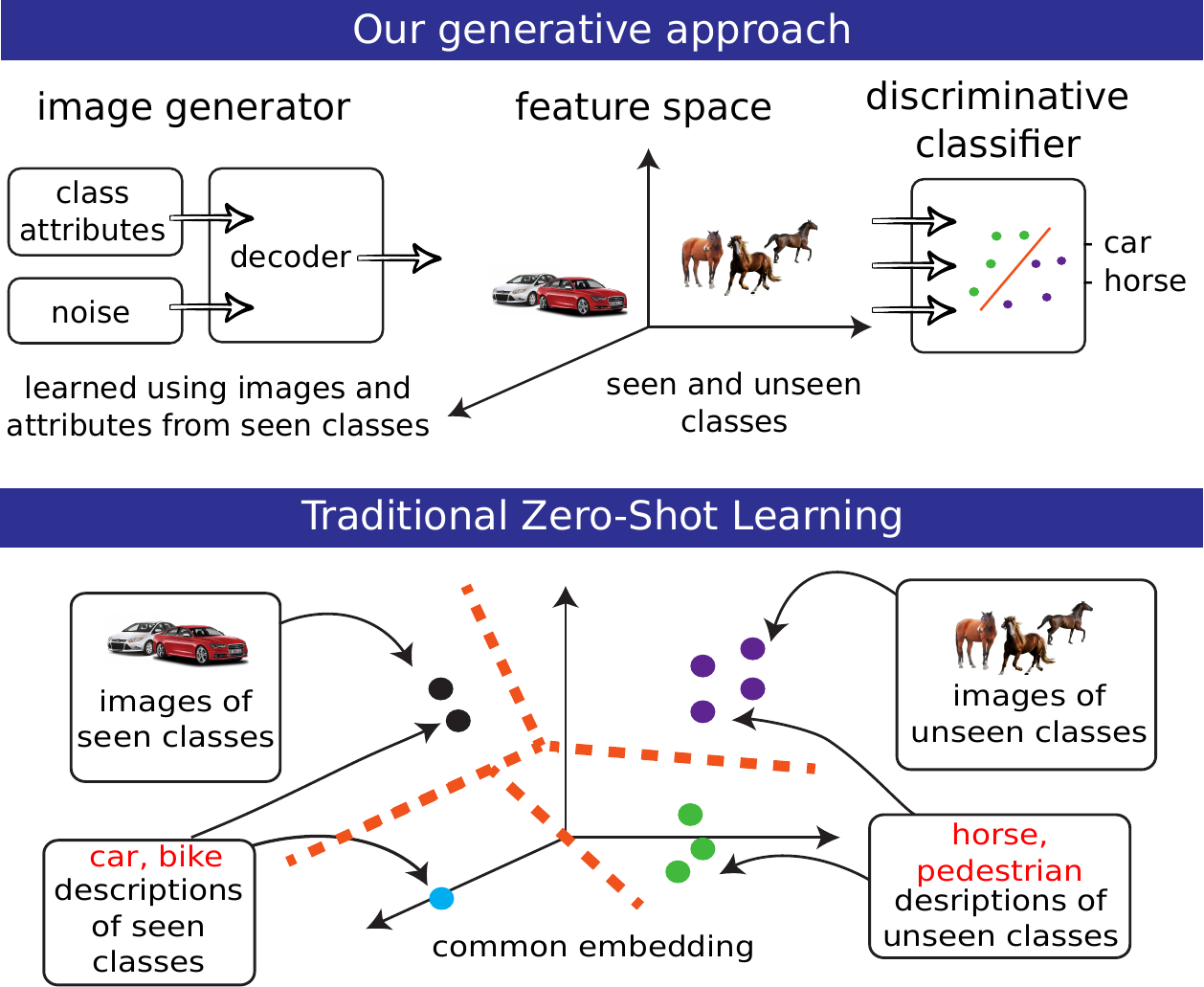}
\caption{Our method consists in i) learning an image feature generator capable of generating artificial image representations from given attributes ii) learning a discriminative classifier from the artificially generated training data.  }
\label{fig:method}
\end{figure}

Zero-Shot Classification (ZSC) \cite{Lampert:2014fs} addresses  classification problems where not all the classes are represented in the training examples.  ZSC can be made possible by defining a high-level description of the categories, relating the new classes ({\em the unseen classes}) to classes for which training examples are available ({\em seen classes}). Learning is usually done by leveraging an intermediate level of representation, the attributes, that provide semantic information about the categories to classify. As pointed out by \cite{romera2015embarrassingly} this paradigm can be compared to how human can identify a new object from a description of it, leveraging similarities between its description and previously learned concepts.  

Recent ZSC algorithms (\eg \cite{Akata:2016kq,Bucher:2016tv}) do the classification by defining a zero-shot prediction function that outputs the class $y$ having the maximum compatibility score with the image $x$: $f(x) = \operatorname*{arg\,max}_y S(x,y)$.  The compatibility function, for its part, is often defined as $S(x, y; W ) = \theta(x)^t W \phi(y)$ where $\theta$ and $\phi$ are two projections and $W$ is a bilinear function relating the two in a common embedding. There are different variants in the recent literature on how the projections or the similarity measure are computed \cite{deng2009imagenet,changpinyo2016synthesized,Frome:2013ux,norouzi2013zero,romera2015embarrassingly,weston2011wsabie,xian2016latent,Zhang:2015vs}, but in all cases the class is chosen as the one maximizing the compatibility score. This embedding and maximal compatibility approach, however, does not exploit, in the learning phase, the information potentially contained in the semantic representation of the unseen categories. The only step where a discriminating capability is exploited is in the final label selection which uses an $\operatorname*{arg\,max}_y$ decision scheme, but not in the setting of the compatibility score itself.

A parallel can be easily done between the aforementioned approaches and {\em generative models} such as defined in the machine learning community. Generative models estimate the joint distribution $p(y,x)$ of images and classes, often by learning the class prior probability $p(y)$ and the class-conditional density $p(x|y)$ separately. However, as it has been observed for a long time \cite{Ulusoy:2005bl}, discriminative approaches trained for predicting directly the class label have better performance than model-based approaches as long as the learning database reliably samples the target distribution.

Despite one can expect discriminative methods to give better performance \cite{Ulusoy:2005bl}, they can't be used directly in the case of ZSC for obvious reasons: as no images are available for some categories, discriminative classifiers cannot be learned out-of-the-box. 

This paper proposes to overcome this difficulty by generating training features for the unseen classes, in such a way that standard discriminative classifiers can be learned (Fig.~\ref{fig:method}).  Generating data for machine learning tasks has been studied in the literature \eg, \cite{Guo:2004dh} or \cite{BaderElDen:2016du} to compensate for imbalanced training sets. Generating novel training examples from the existing ones is also at the heart of the technique called {\em Data Augmentation}, frequently used for training deep neural networks \cite{LeCun:1998hy}. When there is no training data at all for some categories, some underlying parametric representation can be used to generate missing training data, assuming a mapping from the underlying representation to the image space. \cite{Eggert:2015ia} generated images by applying warping and other geometric / photometric transformations to prototypical logo exemplars. A similar idea was also presented in \cite{jaderberg2015reading} for text spotting in images. \cite{Cabrera:2016ix} capture what they call {\em The Gist of a Gesture} by recording human gestures, representing them by a model and use this model to generate a large set of realistic gestures. 

We build in this direction, in the context of ZSC, the underlying representation being some attribute or text based description of the unseen categories, and the transformation from attributes to image features being learned from the examples of the seen classes. A relevant way to learn this transformation is to use generative models such as  {\em denoising auto encoders} \cite{bengio2013generalized} and {\em generative adversarial nets} (GAN) \cite{Goodfellow:2014td} or their variants \cite{Che:2016wq,makhzani2015adversarial}. GANs consist in estimating generative models via an adversarial process simultaneously learning two models,  a generative model that captures the data distribution, and a discriminative model that estimates the probability that a sample came from the training data rather than the generator. The {\em Conditional Generative Adversarial Nets} of \cite{Mirza:2014wi} is a very relevant variant adapted to our problem.  

In addition to the advantage of using discriminative classifiers -- which is expected to give better performance -- our approach, by nature, can address the more realistic task of Generalized Zero-Shot Classification (GZSC). This problem, introduced in \cite{chao2016empirical}, assumes  that both seen and unseen categories are present at test time, making the traditional approaches suffering from bias decision issues.  In contrast, the proposed approach uses (artificial) training examples of both seen and unseen classes during training, avoiding the aforementioned issues. 

Another reason to perform classification inference directly in the visual feature space rather than in an abstract attribute or embedding space is that data are usually more easily separated in the former, especially when using  discriminant deep features that are now commonly available. %Fig.~\ref{fig:attr_feat_tsne} displays a 2D projection of the two kinds of representations and justifies this intuition. It can therefore be expected that a reliable conditional visual feature generator will be able to translate ZSC into a well separated supervised classification problem. 

%\begin{figure}[tb]
%    \centering
%    \subfloat[Attribute space]{{\includegraphics[width=0.21\textwidth]{cub_attr_visu.png} }}%
%    \qquad
 %   \subfloat[Feature space]{{\includegraphics[width=0.21\textwidth]{cub_feat_visu.png} }}%
%    \qquad
 %     \caption{CUB dataset: T-SNE visualization  of feature vectors in (a) attribute space and (b) visual feature space. Data seem visually better separated in visual feature space.}%
 %   \label{fig:attr_feat_tsne}%
%\end{figure}

This paper experimentally validates the proposed strategy on  4 standard Zero-Shot classification datasets (Animals with Attributes (AWA) \cite{Lampert:2014fs}, SUN attributes (SUN) \cite{patterson2012sun}, Apascal\&Ayahoo (aP\&Y) \cite{farhadi2009describing} and Caltech-UCSD Birds-200-2011 (CUB) \cite{wah2011caltech}), and gives insight on how the approach scales on large datasets such as ImageNet \cite{deng2009imagenet}. It shows state-of-the-art performance on all datasets for both ZSC and GZSC.

\section{Approach}

\subsection{Zero shot classification}

As motivated in the introduction, we address in this paper the problem of learning a classifier capable of discriminating between a given set of classes where empirical data is only available for a subset of it, the so-called \emph{seen} classes. In the vocabulary of zero-shot classification, the problem is usually qualified as inductive --- we do not have access to any data from the unseen classes --- as opposed to transductive where the \emph{unseen} data is available but not the associated labels. We do not address in this paper the transductive setting, considering that the availability of target data is a big constraint in practice.  

The learning dataset $\mathcal{D}_s$ is defined by a series of triplets $\left\{\boldmath{x}^s_i, \boldmath{a}^s_i, y^s_i  \right\}_{i=1}^{N_s}$
where $\boldmath{x}^s_i\in \mathcal{X}$ is the raw data (image or features), $y^s_i \in \mathcal{Y}_s$ is the associated class label and $\boldmath{a}^s_i$ is a rich semantic representation of the class (attributes, word vector or text) belonging to $\mathcal{A}_s$. This semantic representation is expected  to i) contain enough information to discriminate between classes by itself, ii) be predictable from raw data and iii) infer unambiguously the class label $y = l(\mathbf{a})$. 

In an inductive ZSC problem, all that is known regarding the new target domain is the set of semantic class representations $\mathcal{A}_u $ of the \emph{unseen} classes. The goal is to use this information and the structure of the semantic representation space to design a classification function $f$ able to predict the class label $\hat{y} = f(\mathbf{x}; \mathcal{A}_u, \mathcal{D}_s)$.
The classification function $f$ is usually parametric and settled by the optimization of an empirical learning criterion.

\subsection{Discriminative approach for ZSC}

In ZSC, the main problem is precisely the fact that no data is available for the unseen classes. The approach taken in this paper is to artificially generate data for the unseen classes given that seen classes and their semantic representations provide enough information to do so, and then apply a discriminative approach to learn the class predictor.

The availability of data for the unseen classes has two main advantages: it can make the classification of seen \emph{and} unseen classes as a single homogeneous process, allowing to address Generalized Zero Shot Classification as a single supervised classification problem; it potentially allows a larger number of unseen classes, which is for instance required for datasets such ImageNet \cite{deng2009imagenet}.

Let $\widehat{\mathcal{D}}_u = \left\{\hat{\boldmath{x}}^u_i, \boldmath{a}^u_i, y^u_i  \right\}_{i=1}^{N_u}$ be a database generated to account for the unseen semantic class representation $\boldmath{a}^u \in \mathcal{A}_u $. The ZSC classification function becomes: $\hat{y} = f_D(\mathbf{x}; \widehat{\mathcal{D}}_u, \mathcal{D}_s)$
and can be used in association with the seen data $\mathcal{D}_s$, to learn a homogeneous supervised problem.

\subsection{Generating unseen data}
\label{sec:gen_mod}

Our generators of unseen data build on the recently proposed approaches for conditional data generation as presented in section~\ref{sec:related_work}. The idea is to learn globally a parametric random generative process $G$ using a differentiable criterion able to compare, as a whole, a target data distribution and a generated one.

\begin{figure}[tb]
\centering
\includegraphics[scale=0.22]{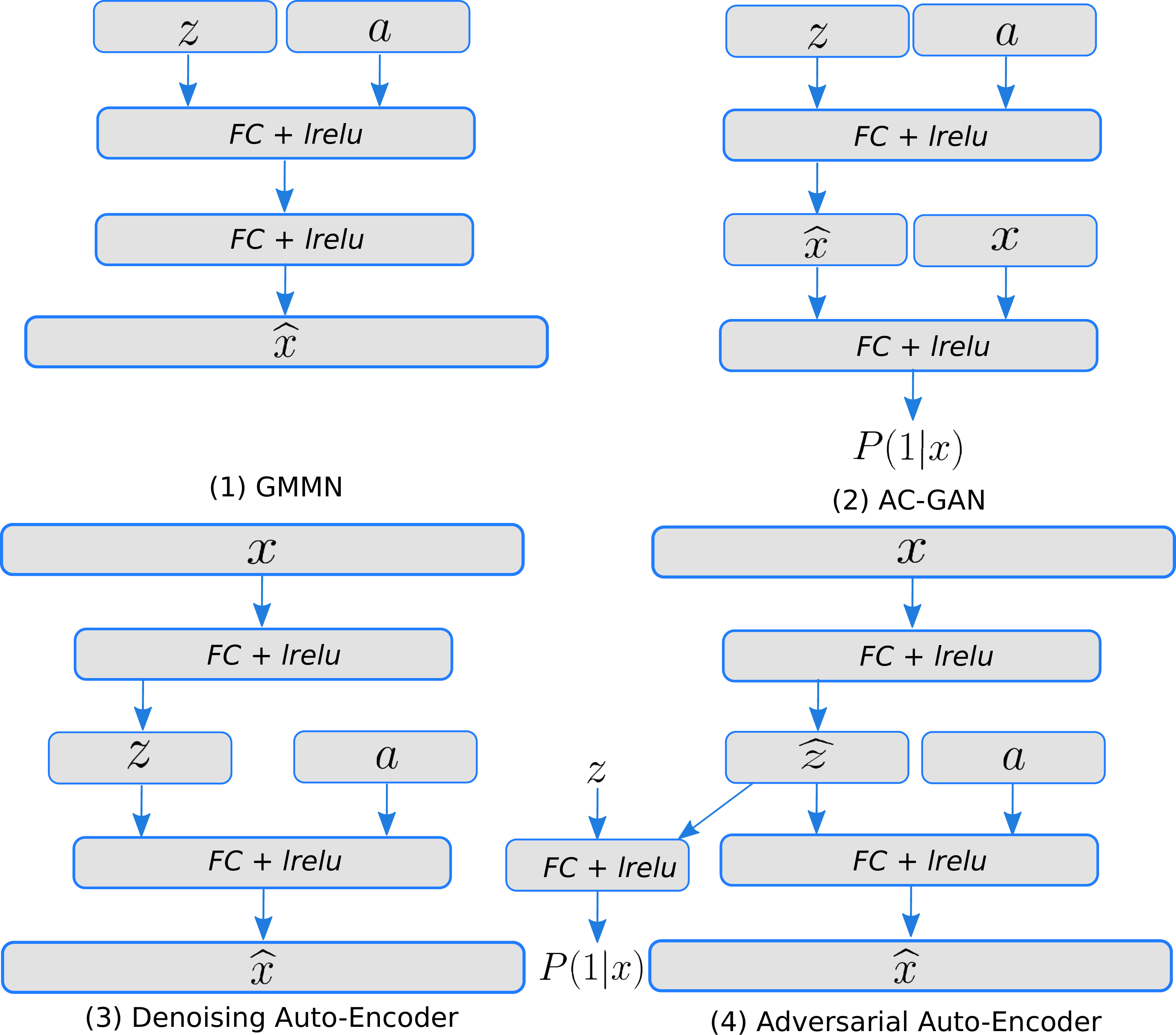}
\caption{Architecture of the different generative models studied.}
\label{fig:gen_model}
\end{figure}

Given  $\mathbf{z}$ a random sample from a fixed multivariate prior distribution, typically uniform or Gaussian, and $\mathbf{w}$ the set of parameters, new sample data consistent with the semantic description $\mathbf{a}$ are generated by applying the function: $\widehat{\boldmath{x}} = G(\mathbf{a},\mathbf{z};\mathbf{w})$. A simple way to generate conditional $\widehat{\boldmath{x}}$ data is to concatenate the semantic representation $\mathbf{a}$ and the random prior $\mathbf{z}$  as the input of a multi-layer network, as shown in Fig.~\ref{fig:gen_model}. 

We now present 4 different strategies to design such a conditional data generator, the functional structure of the generator being common to all the described approaches.

\paragraph{Generative Moment Matching Network}

A first approach is to adapt the Generative Moment Matching Network (GMMN) proposed in~\cite{li2015generative} to conditioning. The generative process will be considered as good if for each semantic description $\mathbf{a}$ two random populations $\mathcal{X}(\mathbf{a})$ from $\mathcal{D}_s$ and $\widehat{\mathcal{X}}(\mathbf{a};\mathbf{w})$ sampled from the generator have low \emph{maximum mean discrepancy} which is a probability divergence measure between two distributions. This divergence can be approximated using a Hilbert kernel based statistics \cite{gretton2012kernel} -- typically a linear combination of Gaussian functions with various widths --- which has the big advantage of being differentiable and may be thus exploited as a machine learning cost. Network parameters $\mathbf{w}$ are then obtained by optimizing the differentiable statistics by stochastic gradient descent, using batches of generated and real data conditioned by the semantic description $\mathbf{a}$.

\paragraph{Conditional Generative adversarial models}

Our second model builds on the principles of the generative adversarial networks (GAN), which is to learn a discrepancy measure between a true and a generated distributions --- the \emph{Discriminator} --- simultaneously with the data generator. One  extension allowing to produce conditional distributions is the AC-GAN \cite{odena2016conditional} (Fig.~\ref{fig:gen_model}) where the generated and the true distributions are compared using a binary classifier, and the quality of the conditional generation is controlled by the performance of this auxiliary task. 

This model bears similarities with the GMMN model, the key difference being that in the GMMN distributions of true and generated data are compared using the kernel based empirical statistics while in the AC-GAN case it is measured by a learned discriminative parametric model.

\paragraph{Denoising Auto-Encoder}
Our third generator relies on the work presented in \cite{bengio2013generalized}, where an encoder/decoder structure is proposed to design a data generator, the latent code playing the role of the random prior $\mathbf{z}$ used to generate the data. A simple extension able to introduce a conditional data generation control has been developed by concatenating the semantic representation $\mathbf{a}$ to the code that is fed to the decoder (Fig.~\ref{fig:gen_model}). 

In practice, this model is learned as a standard auto-encoder, except that i) some noise is added to the input and ii)  the semantic representation $\mathbf{a}$ is concatenated to the code in the hidden layer. For generating novel examples, only the decoder part, \ie the head of the network using $\mathbf{z}$ and $\mathbf{a}$ as input to produce $\widehat{\mathbf{x}}$ is used.

%The other family of data generator exploits a game theoretic optimization approach now referred to \emph{adversarial} networks. Compared to GMMN, it introduces an extra objective --- the discrimination between synthesized ans real data --- which is jointly optimized with the data generation itself. 

\paragraph{Adversarial Auto-Encoder}
Our fourth generator is inspired by \cite{makhzani2015adversarial}, which  is an extension of the denoising auto-encoder. It introduces an adversarial criterion to control the latent code produced by the encoder part, so that the code distribution matches a fixed prior distribution. This extra constraint is expected to ensure that all parts of the sampling prior space will produce meaningful data.

During training, both the auto-encoder and the discriminator are learned simultaneously. For generating novel examples, as for the denoising auto-encoder, only the decoder part is used. 

% \subsection{Conditional data generation for ZSC}
% (Cross validation on hyper parameters)
% (Deep features are expected to be already well separated for the target data and categories.)

% \subsection{Evaluation of ZSC}
% \textcolor{red}{Here or in the evaluation section ?}
% Extend the scope of classification.

\subsection{Implementing the generators}

We implemented our 4 generative models with neural networks, whose architectures are illustrated Fig.~\ref{fig:gen_model}. Hidden layers are fully connected (FC) with leaky-relu non-linearity \cite{maas2013rectifier} (leakage coefficient of 0.2). For the models using a classifier (the AC-GAN and the Adversarial auto-encoder), the classifier is a linear classifier (fully connected layer + Softmax activation function). The loss used to measure the quality of the reconstruction in the two auto-encoders is the L2 norm.

Regarding how to sample the noise $\mathbf{z}$, we did not observe any difference between sampling it from a Gaussian distribution or from a uniform distribution.

\section{Experiments}

In this section, after presenting the datasets and the experimental settings, we start by comparing the different generative models described in the previous section. We then show how our approach can be used for the Generalized Zero-shot Classification Task, which is one of the key contributions of the paper, provide some experiments on a large scale zero shot classification task, and finally compare our approach with state-of-the art Zero-Shot approaches on the regular Zero-shot Classification Task.

\subsection{Datasets and Settings}
\label{sec:sett}

% \begin{table}[tb]
%     \scriptsize
% 	\centering
% 	\caption{Dataset statistics. SUN dataset is used with two distinct types of split (see \ref{sec:sett} for explanations).}
% 	\label{statsdata}
% 	\begin{tabular}{|l|c|c|c|c|}
% 		\hline
% 		Dataset &\#Seen classes & \#Unseen classes&\#Instances& \#Attr./\#w2c \\\hline\hline
% 		aP\&Y  & 20 & 12 & 15,339 & 64 \\\hline
% 		AwA  & 40 & 10 & 30,475 & 85 \\\hline
% 		CUB  & 150 & 50 & 11,788 & 312 \\\hline
% 		SUN  & 707 (645/646) & 10 (72/71) & 14,340 & 102 \\\hline
% 		ImageNet  & 1000 & 20842 & 14,197,122 & 1000 \\\hline
% 	\end{tabular}
% \end{table}

A first experimental evaluation is done on 4 standard ZSC datasets: Animals with Attributes (AWA) \cite{Lampert:2014fs}, SUN attributes (SUN) \cite{patterson2012sun}, Apascal\&Ayahoo (aP\&Y) \cite{farhadi2009describing} and Caltech-UCSD Birds-200-2011 (CUB) \cite{wah2011caltech} 
%(see table \ref{statsdata} for statistics)
. These benchmarks exhibit a great diversity of concepts; SUN and CUB are for fine-Grained categorization, and  include respectively  birds and scenes images; AwA contains images of animals from 50 different categories; finally,  aP\&Y has broader concepts, from cars to animals. For each dataset, attributes descriptions are given, either at the class level or at image level.  aP\&Y, CUB and SUN have per image binary attributes that we average to produce per class real valued representations. In order to make comparisons with other works, we follow the same training/testing splits for AwA \cite{Lampert:2014fs}, CUB \cite{Akata:2015tv} and aP\&Y \cite{farhadi2009describing}. For SUN we experiment two different settings: one with 10 unseen classes as in \cite{Jayaraman:2014wq}, a second, more competitive, with ten different folds randomly chosen and averaged, as proposed by \cite{changpinyo2016synthesized} (72/71 splits).

Image features are computed using two deep networks,  the VGG-VeryDeep-19  \cite{simonyan2014very} and the GoogLeNet \cite{szegedy2014going} networks. For the VGG-19 we use the 4,096-dim top-layer hidden unit activations (fc7) while for the GoogLeNet we use the 1,024-dim top-layer pooling units. We keep the weights learned on ImageNet fixed \ie, we don't apply any fine-tuning.

The classifiers are obtained by adding a standard Fully Connected with Softmax layer to the pre-trained networks. We purposively chose a simple classifier to better observe the behavior of the generators. In all our experiments we generated 500 artificial image features by class,  which we consider to be a reasonable trade-off between accuracy and training time; we have not observed any significant improvement when adding more images.

Each architecture has its own set of hyper-parameters (typically the number of units per layer, the number of hidden layers, the learning rate, \etc). They are obtained trough a 'Zero-shot' cross-validation procedure. In this procedure, 20\% of the seen classes are considered as unseen (hence used as validation set), allowing to choose the hyper-parameters maximizing the accuracy on this so-obtained validation set.
In practice, typical values for the number of neurons (resp. the number of hidden layers) are in the range  of [500-2000] (resp. 1 or 2).

Model parameters are initialized according to a centered Gaussian distribution ($\sigma= 0.02$). They are optimized with the Adam solver \cite{kingma2014adam} with a cross-validated learning rate (typically of $10^{-4}$), using mini-batches of size 128 except for the GMMN where each batch contains all the training images of one class, to make the estimation of the statistics more reliable. In order to avoid over-fitting, we used dropout  \cite{srivastava2014dropout} at every layer (probability of drop of 0.2 for the inputs layers  and of 0.5 for the hidden layers). Input data (both image features and w2c vectors) are scaled to [0,1] by applying an affine transformation. With the TensorFlow framework  \cite{tensorflow2015-whitepaper} running on a Nvidia Titan X pascal GPU, the learning stage takes around 10 minutes for a given set of hyper-parameters. Our code will be made publicly available.
Our code will be made publicly available.\footnote{https://github.com/maximebucher}

\subsection{Comparing the different generative models}
\label{sec:generative_models_comparison}

%  \begin{table*}[tb]
%              	\centering
%              	\footnotesize 
%              	\caption{Zero-Shot classification accuracy (mean$\pm$std) on the validation set, for the 4 generative models.}
%              	\label{gen_score}
%              	\begin{tabular}{|l|l|l|l|l|l|l|}
%              		\hline
%              		Model  & {aP\&Y} & {AwA} & {CUB} & {SUN} & Avg \\  \hline	
%              		Denoising Auto-encoder \cite{bengio2013generalized} & {61.99$\pm1.19$} & {66.36$\pm0.68$} & {42.83$\pm0.35$} & {82.50$\pm0.42$} & 63.42 \\
%              		AC-GAN \cite{odena2016conditional} & {55.23$\pm2.01$}  & {66.00$\pm0.66$}  & {44.55$\pm0.82$} & {83.5$\pm0.74$} & 62.32\\
%              		Adversarial Auto-encoder \cite{makhzani2015adversarial} & {59.50$\pm2.52$} & \textbf{68.42$\bf{\pm0.13}$} & {49.78$\pm0.83$} & {83.71$\pm0.88$} & 65.35\\
%              		GMMN \cite{li2015generative} & \textbf{65.85$\bf{\pm1.36}$} & 67.04$\pm0.12$  & \textbf{52.35$\bf{\pm0.92}$}  & \textbf{84.00$\bf{\pm0.53}$}  & \textbf{67.31} \\
%              		\hline
%              	\end{tabular}
%              \end{table*}
             
 \begin{table}[tb]
             	\centering
             	\footnotesize 
             	\caption{Zero-Shot classification accuracy (mean) on the validation set, for the 4 generative models.}
             	\label{gen_score}
             	\begin{tabular}{l|llll|l}
             	
             		\textbf{Model}  & \textbf{aP\&Y} & \textbf{AwA} & \textbf{CUB} & \textbf{SUN} & \textbf{Avg} \\  \hline	
             		Denois. Auto-encod. \cite{bengio2013generalized} & {62.0} & {66.4} & {42.8} & {82.5} & 63.4 \\
             		AC-GAN \cite{odena2016conditional} & {55.2}  & {66.0}  & {44.6} & {83.5} & 62.3\\
             		Adv. Auto-encod. \cite{makhzani2015adversarial} & {59.5} & \textbf{68.4} & {49.8} & {83.7} & 65.3\\
             		GMMN \cite{li2015generative} & \textbf{65.9} & 67.0  & \textbf{52.4}  & \textbf{84.0}  & \textbf{67.3} \\
             	
             	\end{tabular}
             \end{table}

Our first round of experiments consists in comparing the performance of the 4 generative models described in Section~\ref{sec:gen_mod},  on the regular Zero-shot classification task. Our intention is to select the best one for further experiments. Performance on the validation set is reported Table \ref{gen_score}. We can see that the GMMN model outperforms the 3 others on average, with a noticeable 5\% improvement on aP\&Y. Its optimization is also computationally more stable than the adversarial versions. We consequently chose this generator for the following.  

% \begin{figure}[tb]
% \centering
% \includegraphics[scale=0.22]{visu_gen.png}
% \caption{AwA dataset: T-SNE visualization of artificial image representations  generated by the GMMN, AC-GAN, ADV AE (Adversarial Auto-encoder), AE (Denoising Auto-encoder) and TRUE (real data). Best viewed on a computer screen with strong zoom factor.}
% \label{fig:visu_gen}
% \end{figure}
% In order to have a better understanding of the behavior of the 4 models, we visualized the generated data, after projecting them in 2D with a T-SNE projection \cite{maaten2008visualizing}. The projections are given Fig.~\ref{fig:visu_gen}). For each model, 10 images per class are generated and compared to 10 real images taken from unseen classes. As we can see, the GMMN model produced clusters that are, on average, very  similar to the real data clusters. 

We explain the superiority of the GMMN model by the fact it aligns the distributions by using an explicit model of the divergence of the distributions while the adversarial autoencoder and the AC-GAN have to learn it. For its part, the denoising autoencoder doesn't have any guaranty that the distributions are aligned, explaining its weak performance compared to the 3 other generators.  

\subsection{Generalized Zero-Shot Classification task}
\begin{table}[]
                       	\centering
                       	\footnotesize
                       	\caption{Generalized Zero-Shot classification accuracy on AWA. Image features are obtained with the GoogLeNet. \cite{szegedy2014going} CNN.}
                       	\label{generalized_awa}
                        \setlength\tabcolsep{4.5pt} 
                       	\begin{tabular}{l|lllllllll}
                       		
                       		& \multicolumn{4}{l}{\textbf{AwA}} \\ 
                       		\textbf{Method} & $ \bm{u \rightarrow u}$    &   $\bm{s \rightarrow s}$   &   $\bm{u \rightarrow a}$   &   $\bm{s \rightarrow a}$    \\ \hline
                       		Lampert \etal \cite{Lampert:2014fs}$^{dap}$ &   51.1  &  78.5   &  2.4   &   77.9   \\ 
                       		Lampert \etal \cite{Lampert:2014fs}$^{iap}$&   56.3  &  77.3   &   1.7  &   76.8  \\ 
                       		Norouzi \etal \cite{norouzi2013zero} & 63.7  &  76.9   &   9.5  &   75.9  \\ 
                       		Changpinyo \etal \cite{changpinyo2016synthesized}$^{o-vs-o}$  & 70.1    & 67.3    & 0.3    & 67.3   \\ 
                       		Changpinyo \etal \cite{changpinyo2016synthesized}$^{struct}$ & 73.4    & 81.0    & 0.4    & 81.0   \\ \hline 
                       		Ours &  \textbf{77.12}   &   \textbf{82.73}  &  \textbf{32.32}   & \textbf{ 81.32 }    \\ \hline 
                             Ours. (VGG-19) &  87.78   &   85.61  &  38.21   &  83.14       \\ 
                       	\end{tabular}
                       \end{table}

\begin{table}[]
                    	\centering
                    	\footnotesize
                    	\caption{Generalized Zero-Shot classification accuracy on CUB. Image features are obtained with the GoogLeNet \cite{szegedy2014going} CNN.}
                    	\label{generalized_cub}
                        \setlength\tabcolsep{4.5pt} 
                    		\begin{tabular}{l|lllllllll}
                    	& \multicolumn{4}{l}{\textbf{CUB}} \\ 
                       		\textbf{Method} & $ \bm{u \rightarrow u}$    &   $\bm{s \rightarrow s}$   &   $\bm{u \rightarrow a}$   &   $\bm{s \rightarrow a}$    \\ \hline
                    		Lampert \etal \cite{Lampert:2014fs}$^{dap}$  & 38.8 &   56.0    &  4.0   & 55.1   \\ 
                    		Lampert \etal \cite{Lampert:2014fs}$^{iap}$&  36.5 &   69.6    &  1.0   & 69.4   \\ 
                    		Norouzi \etal \cite{norouzi2013zero} & 35.8 &   70.5    &  1.8   & 69.9    \\ 
                    		Changpinyo \etal \cite{changpinyo2016synthesized}$^{o-vs-o}$  & 53.0 &    67.2    & 8.4    &  66.5   \\ 
                    		Changpinyo \etal \cite{changpinyo2016synthesized}$^{struct}$  & 54.4 &    \textbf{73.0}    & 13.2    & 72.0     \\ \hline 
                    		Ours.   &  \textbf{60.10}   &  72.38 & \textbf{26.87} &  \textbf{72.00}    \\ \hline 
                           Ours. (VGG-19)  &  59.70   &  71.21 & 20.12 &  69.45    \\ 
                    	\end{tabular}
                    \end{table}

In this section, we follow the Generalized Zero-Shot Learning (GZSC) protocol introduced by Chao \etal \cite{chao2016empirical}. In this protocol, test data are from any classes, seen or unseen. This task is more realistic and harder, as the number of class candidates is larger.

We follow the notations of \cite{chao2016empirical}, \ie \\
$u \rightarrow u$: test images from unseen classes, labels of unseen classes  (conventional ZSC)\\
$s \rightarrow s$: test images from seen classes, labels of seen classes (multi-class classication for seen classes)\\
$u \rightarrow a$: test images from unseen classes, labels of seen and unseen classes (GZSC)\\
$s \rightarrow a$: test images from seen classes, labels of seen and unseen classes (GZSC)

In the first two cases, only the seen/unseen classes are used in the training phase. In the last two cases, the classifier is learned with training data combining images generated for all classes (seen and not seen). 

Most of the recent ZSC works \eg, \cite{Akata:2015tv,bucher2016hard,Bucher:2016tv,romera2015embarrassingly} are focused on improving the embedding or the scoring function. However, \cite{chao2016empirical} has shown that this type of approach is unpractical with GZSC. Indeed the scoring function is in this case biased toward seen classes, leading to very low accuracy on the unseen classes. This can be seen on Table \ref{generalized_awa} and \ref{generalized_cub} ($u \rightarrow a$ column), where the accuracy drops significantly compared to regular ZSC performance. 
The data distribution of the ZSC datasets are strongly subject to this bias, as unseen classes are very similar to seen classes both in terms of visual appearance and attribute description. When seen and unseen classes are candidates, it becomes much harder to distinguish between them. For example, the horse (seen)  and the zebra classes (unseen) of the AwA dataset cannot be distinguished by standard ZSC methods. 

As we can see on Table \ref{generalized_awa} and \ref{generalized_cub}, our generative approach outperforms any other previous approach. In the hardest case, $u \rightarrow a$, it gives the accuracy of 30\% (resp.  10\%) higher than state-of-the-art approaches on the AwA (resp. CUB) dataset. It can be easily explained by the fact that it doesn't suffer from the scoring function problem we mentioned, as the Softmax classifier is learned to discriminate both seen and unseen classes, offering a decisive solution to the bias problem.

\subsection{Large Scale Zero-Shot Classification}
\begin{table}[]
                     \centering
                     \footnotesize 
                     \caption{Zero-shot and Generalized ZSC on ImageNet.}
                     \label{imagenet}
                     \setlength\tabcolsep{4.5pt} 
                     \begin{tabular}{l|l|lllll}
                    
                      &  &               & \textbf{Flat}                   & \textbf{Hit}                    & \textbf{@K}                      &  \\
   \textbf{Scenario}                              & \textbf{Method}    & \multicolumn{1}{l}{\textbf{1}} & \multicolumn{1}{l}{\textbf{2}} & \multicolumn{1}{l}{\textbf{5}} & \multicolumn{1}{l}{\textbf{10}} & \textbf{20}        \\ \hline
                  \textbf{2-hop}         & Frome  \cite{Frome:2013ux}  & \multicolumn{1}{l}{6.0} & \multicolumn{1}{l}{10.0}  & \multicolumn{1}{l}{18.1}  & \multicolumn{1}{l}{26.4}   &     36.4      \\ 
                                & Norouzi  \cite{norouzi2013zero}   & \multicolumn{1}{l}{9.4}  & \multicolumn{1}{l}{15.1}  & \multicolumn{1}{l}{24.7}  & \multicolumn{1}{l}{32.7}   &    41.8       \\  
                                & Changpinyo  \cite{changpinyo2016synthesized}    & \multicolumn{1}{l}{10.5}  & \multicolumn{1}{l}{16.7}  & \multicolumn{1}{l}{28.6}  & \multicolumn{1}{l}{40.1}   &      52.0    \\ \cline{2-7}
                                & Ours.   & \multicolumn{1}{l}{\textbf{13.05}}  & \multicolumn{1}{l}{\textbf{21.52}}  & \multicolumn{1}{l}{\textbf{33.71}}  & \multicolumn{1}{l}{\textbf{43.91}}   &  \textbf{57.31  }      \\ \hline
                               
                \textbf{2-hop}      & Frome \cite{Frome:2013ux}  & \multicolumn{1}{l}{0.8}  & \multicolumn{1}{l}{2.7}  & \multicolumn{1}{l}{7.9}  & \multicolumn{1}{l}{14.2}   &     22.7      \\ 
                     \textbf{(+1K)}           & Norouzi \cite{norouzi2013zero}   & \multicolumn{1}{l}{0.3}  & \multicolumn{1}{l}{7.1}  & \multicolumn{1}{l}{17.2}  & \multicolumn{1}{l}{24.9}   &    33.5       \\ \cline{2-7}
                                
                                & Ours.   & \multicolumn{1}{l}{\textbf{4.93}}  & \multicolumn{1}{l}{\textbf{13.02}}  & \multicolumn{1}{l}{\textbf{20.81}}  & \multicolumn{1}{l}{\textbf{31.48}}   &   \textbf{45.31 }      \\ \hline \hline
              \textbf{3-hop}             & Frome  \cite{Frome:2013ux}  & \multicolumn{1}{l}{1.7}  & \multicolumn{1}{l}{2.9}  & \multicolumn{1}{l}{5.3}  & \multicolumn{1}{l}{8.2}   &   12.5        \\ 
                                & Norouzi  \cite{norouzi2013zero}   & \multicolumn{1}{l}{2.7}  & \multicolumn{1}{l}{4.4}  & \multicolumn{1}{l}{7.8}  & \multicolumn{1}{l}{11.5}   &     16.1      \\  
                                & Changpinyo  \cite{changpinyo2016synthesized}    & \multicolumn{1}{l}{2.9}  & \multicolumn{1}{l}{4.9}  & \multicolumn{1}{l}{9.2}  & \multicolumn{1}{l}{14.2}   &   20.9        \\ \cline{2-7}
                                & Ours.   & \multicolumn{1}{l}{\textbf{3.58}}  & \multicolumn{1}{l}{\textbf{5.97}}  & \multicolumn{1}{l}{\textbf{11.03}}  & \multicolumn{1}{l}{\textbf{16.51}}   &  \textbf{23.88 }       \\ \hline         
              \textbf{3-hop}        & Frome  \cite{Frome:2013ux}  & \multicolumn{1}{l}{0.5}  & \multicolumn{1}{l}{1.4}  & \multicolumn{1}{l}{3.4}  & \multicolumn{1}{l}{5.9}   &    9.7       \\ 
                    \textbf{(+1K)}            & Norouzi  \cite{norouzi2013zero}   & \multicolumn{1}{l}{0.2}  & \multicolumn{1}{l}{2.4}  & \multicolumn{1}{l}{5.9}  & \multicolumn{1}{l}{9.7}   &    14.3       \\ \cline{2-7}
                               
                                & Ours.   & \multicolumn{1}{l}{\textbf{1.99}}  & \multicolumn{1}{l}{\textbf{4.01}}  & \multicolumn{1}{l}{\textbf{6.74}}  & \multicolumn{1}{l}{\textbf{11.72}}   &   \textbf{16.34}       \\ \hline  \hline
              \textbf{All}              & Frome  \cite{Frome:2013ux}  & \multicolumn{1}{l}{0.8}  & \multicolumn{1}{l}{1.4}  & \multicolumn{1}{l}{2.5}  & \multicolumn{1}{l}{3.9}   &    6.0       \\ 
                                & Norouzi  \cite{norouzi2013zero}   & \multicolumn{1}{l}{1.4}  & \multicolumn{1}{l}{2.2}  & \multicolumn{1}{l}{3.9}  & \multicolumn{1}{l}{5.8}   &     8.3      \\ 
                                & Changpinyo  \cite{changpinyo2016synthesized}    & \multicolumn{1}{l}{1.5}  & \multicolumn{1}{l}{2.4}  & \multicolumn{1}{l}{4.5}  & \multicolumn{1}{l}{7.1}   &     10.9      \\ \cline{2-7}  
                                & Ours.   & \multicolumn{1}{l}{\textbf{1.90}}  & \multicolumn{1}{l}{\textbf{3.03}}  & \multicolumn{1}{l}{\textbf{5.67}}  & \multicolumn{1}{l}{\textbf{8.31}}   &   \textbf{13.14}       \\ \hline
               \textbf{All}        & Frome  \cite{Frome:2013ux}  & \multicolumn{1}{l}{0.3}  & \multicolumn{1}{l}{0.8}  & \multicolumn{1}{l}{1.9}  & \multicolumn{1}{l}{3.2}   &    5.3       \\ 
                      \textbf{(+1K)}                                & Norouzi  \cite{norouzi2013zero}   & \multicolumn{1}{l}{0.2}  & \multicolumn{1}{l}{1.2}  & \multicolumn{1}{l}{3.0}  & \multicolumn{1}{l}{5.0}   &     7.5      \\ \cline{2-7}                                        
                                                      & Ours.   & \multicolumn{1}{l}{\textbf{1.03}}  & \multicolumn{1}{l}{\textbf{1.93}}  & \multicolumn{1}{l}{\textbf{4.98}}  & \multicolumn{1}{l}{\textbf{6.23}}   &   \textbf{10.26}       \\        
             
                     \end{tabular}
                     \end{table}

We compared our approach with state-of-the-art methods on a large-scale Zero-Shot classification task.  These experiences mirror those presented in \cite{Frome:2013ux}: 1000 classes from those of the ImageNet 2012 1K set \cite{russakovsky2015imagenet} are chosen for training (seen classes) while 20.345 others are considered to be unseen classes with no image available.  %ImageNet \cite{deng2009imagenet} is a large dataset with more than 21000 classes.
Image features are computed with the GoogLeNet network \cite{szegedy2014going}.

In contrast with ZSC datasets, no attributes are provided for defining unseen classes. We represent those categories using a skip-gram language model \cite{mikolov2013distributed}. This model is learned on a dump of the Wikipedia corpus ($\approx$3 billion words). Skip-gram is a language model learned to predict context from words. The neural network has 1 input layer, 1 hidden layer and 1 output layer having the size of the vocabulary (same size as the input layer). The hidden layer has 500 neurons in our implementation. In the literature, the hidden layer has been reported to be an interesting embedding space for representing word. Consequently, We use this hidden layer to describe each class label by embedding the class name into this 500-dimensional space. Some classes cannot be represented as their name is not contained in the vocabulary established by parsing the Wikipedia corpus. Such classes are ignored, bringing the number of classes from 20,842 to 20,345 classes. For fair comparison, we take the same language model as \cite{changpinyo2016synthesized} with the same classes excluded.

As in \cite{changpinyo2016synthesized,Frome:2013ux} our model is evaluated on three different scenarios, with an increasing number of unseen classes:  i) 2-hop: 1,509 classes ii) 3-hop: 7,678 classes, iii) All: all unseen categories. 

For this task we use the Flat-Hit@K metric, the percentage of test images for which the model returns the true labels in the top K prediction scores.

Table~\ref{imagenet} summarizes the performance on the 3 hops. As one can see, our model gets state-of the art performance for each configuration. As it can be observed from these experiments, our generative model is very suitable for this large scale GZSC problem \eg, our approach improves by 5\% best competitors for the Flat-Hit 1 metric on the 2-hop scenario.

\subsection{Classical Zero-Shot Classification task}  

In this last section, we follow the protocol of the standard ZSC task: during training, only data from seen classes are available while at test time new images (from unseen classes only) have to be assigned to one of the unseen classes.

As explained in the introduction, the recent ZSC literature \cite{Akata:2015tv,bucher2016hard,Bucher:2016tv,romera2015embarrassingly} mostly focuses on developing a good embedding for comparing attributes and images. One of our motivations for generating training images was to make the training of discriminative classifiers possible, assuming it would result in better performance. This section aims at validating this hypothesis on the regular ZSC task. 

Table \ref{acczstab} summarizes our experiments, reporting the accuracy obtained by state of the art methods on the 4 ZSC datasets, with 2 different deep image features. Each entry is the mean/standard deviation computed on 5 different runs. 

With the VGG network, our method give above state-of-the-art performance on each dataset, with a noticeable improvement of more than 15\% on CUB. On the SUN dataset, Changpinyo \etal \cite{changpinyo2016synthesized}'s seems to give better performance but used the MIT Places dataset to learn the features. It has been recently pointed out in sec. 5.1 of Xiang et al. \cite{xian2017zero}  that this database "intersects with both training and test classes of SUN”, which could explain their better results compared to ours.

\begin{table}[]
           \centering
           \footnotesize 
           \caption{Zero-shot classification accuracy (mean$\pm$std) on 5 runs. We report results with VGG-19 and GoogLeNet features. SUN dataset is evaluated on 2 different splits (see \ref{sec:sett}). *  \cite{changpinyo2016synthesized} features extracted from an MIT Places\cite{zhou2014learning} pre-trained model.}
           \label{acczstab}
           \setlength\tabcolsep{4.5pt} 
           \begin{tabular}{c|l|llll}
          
           \textbf{Feat.} & \textbf{Method}  & \textbf{aP\&Y} & \textbf{AwA} & \textbf{CUB} & \textbf{SUN}  \\  \hline
          
           & Lampert \etal \cite{Lampert:2014fs} & - & {60.5} & {39.1} & {-/44.5}\\
           & Akata \etal \cite{Akata:2015tv} & - & {66.7} & {50.1} & {-/-}\\
           
           & Changpinyo \etal \cite{changpinyo2016synthesized} & - & {72.9} & {54.7} & \textbf{90.0}/\textbf{62.8}*\\
            \rot{\rlap{\small \textbf{GoogLe}}}\rot{\rlap{\textbf{Net}\cite{szegedy2014going}}}
           & Xian \etal \cite{xian2016latent} & - & {71.9} & {45.5} & {-}\\
           \hline 
           & Ours. & \textbf{ 55.34} & \textbf{ 77.12} & \textbf{60.10} & \textbf{85.50}/\textbf{56.41}\\ 
   
           \hline \hline
           &Lampert \etal \cite{Lampert:2014fs} & 38.16 & 57.23 & - & 72.00/- \\
           &Romera-Paredes \cite{romera2015embarrassingly}  & 24.22 & 75.32 & - & 82.10/- \\
           &Zhang \etal \cite{Zhang:2015vs}  & 46.23 & 76.33 & 30.41 & 82.50/-\\
           &Zhang \etal \cite{zhang2016zero}  & 50.35 & { 80.46} & 42.11 & 83.83/-\\
           &Wang \etal \cite{wang2016zero} & - & 78.3 & 48.6 & -/- \\
           & Bucher \etal \cite{Bucher:2016tv} & 53.15 & { 77.32} & 43.29 & 84.41/-\\ 
           \rot{\rlap{\small  \textbf{VGG-VeryDeep}}}\rot{\rlap{~~~~~~~\cite{simonyan2014very}}}
           & Bucher \etal \cite{bucher2016hard} &{56.77} & {86.55} & {45.87} & {86.21/-}\\ 
           \hline 
           & Ours. & \textbf{{57.19}} & \textbf{{87.78}} & \textbf{{59.70}} & \textbf{{88.01/-}}\\ 
          
           \end{tabular}
           \end{table}
\section{Conclusions}

This paper introduces a novel way to address Zero-Shot Classification and Generalized Zero-Shot Classification tasks by learning a conditional generator from seen data and  generating artificial training examples for the categories without exemplars, turning ZSC into a standard supervised learning problem. This novel formulation addresses the two main limitation of previous ZSC method \ie, their intrinsic bias for  Generalized Zero-Shot Classification tasks and their limitations in using discriminative classifiers in the deep image feature space. Our experiments with 4 generative models and 5 datasets experimentally validate the approach and give state-of-the-art performance. 

%We explain the overall good performance by the use of discriminative classifiers, and by directly discriminating in the image feature space.

\section*{Acknowledgement}
M.Bucher was in part supported by R\'{e}gion Normandie.
\bibliographystyle{plain}

{\small
\bibliographystyle{ieee}
\bibliography{0-bibfile.bib}

\begin{thebibliography}{10}

\bibitem{Akata:2016kq}
Zeynep Akata, Mateusz Malinowski, Mario Fritz, and Bernt Schiele.
\newblock {Multi-cue Zero-Shot Learning with Strong Supervision}.
\newblock In {\em 2016 IEEE Conference on Computer Vision and Pattern
  Recognition (CVPR}, pages 59--68. IEEE, June 2016.

\bibitem{Akata:2015tv}
Zeynep Akata, Scott Reed, Daniel Walter, Honglak Lee, and Bernt Schiele.
\newblock {Evaluation of Output Embeddings for Fine-Grained Image
  Classification}.
\newblock In {\em IEEE International Conference on Computer Vision and Pattern
  Recognition (CVPR)}, 2015.

\bibitem{BaderElDen:2016du}
Mohamed~Bahy Bader-El-Den, Eleman Teitei, and Mo~Adda.
\newblock {Hierarchical classification for dealing with the Class imbalance
  problem.}
\newblock {\em IJCNN}, 2016.

\bibitem{bengio2013generalized}
Yoshua Bengio, Li~Yao, Guillaume Alain, and Pascal Vincent.
\newblock Generalized denoising auto-encoders as generative models.
\newblock In {\em Advances in Neural Information Processing Systems}, pages
  899--907, 2013.

\bibitem{Bucher:2016tv}
M~Bucher, S~Herbin, and F~Jurie.
\newblock {Improving semantic embedding consistency by metric learning for
  zero-shot classiffication}.
\newblock In {\em European Conference on Computer Vision}, 2016.

\bibitem{bucher2016hard}
Maxime Bucher, St{\'e}phane Herbin, and Fr{\'e}d{\'e}ric Jurie.
\newblock Hard negative mining for metric learning based zero-shot
  classification.
\newblock In {\em Computer Vision--ECCV 2016 Workshops}, pages 524--531.
  Springer, 2016.

\bibitem{Cabrera:2016ix}
Maria~E Cabrera and Juan~P Wachs.
\newblock {Embodied gesture learning from one-shot}.
\newblock In {\em 2016 25th IEEE International Symposium on Robot and Human
  Interactive Communication (RO-MAN}, pages 1092--1097. IEEE, 2016.

\bibitem{changpinyo2016synthesized}
Soravit Changpinyo, Wei-Lun Chao, Boqing Gong, and Fei Sha.
\newblock Synthesized classifiers for zero-shot learning.
\newblock In {\em Computer Vision and Pattern Recognition (CVPR), 2016 IEEE
  Conference on}, pages 5327--5336. IEEE, 2016.

\bibitem{chao2016empirical}
Wei-Lun Chao, Soravit Changpinyo, Boqing Gong, and Fei Sha.
\newblock An empirical study and analysis of generalized zero-shot learning for
  object recognition in the wild.
\newblock In {\em European Conference on Computer Vision}, pages 52--68.
  Springer, 2016.

\bibitem{Che:2016wq}
Tong Che, Yanran Li, Athul~Paul Jacob, Yoshua Bengio, and Wenjie Li.
\newblock {Mode Regularized Generative Adversarial Networks}.
\newblock {\em arXiv}, December 2016.

\bibitem{deng2009imagenet}
Jia Deng, Wei Dong, Richard Socher, Li-Jia Li, Kai Li, and Li~Fei-Fei.
\newblock Imagenet: A large-scale hierarchical image database.
\newblock In {\em Computer Vision and Pattern Recognition, 2009. CVPR 2009.
  IEEE Conference on}, pages 248--255. IEEE, 2009.

\bibitem{Eggert:2015ia}
Christian Eggert, Anton Winschel, and Rainer Lienhart.
\newblock {On the Benefit of Synthetic Data for Company Logo Detection.}
\newblock In {\em ACM Multimedia}, 2015.

\bibitem{tensorflow2015-whitepaper}
Mart\'{\i}n~Abadi et~al.
\newblock {TensorFlow}: Large-scale machine learning on heterogeneous systems,
  2015.
\newblock Software available from tensorflow.org.

\bibitem{farhadi2009describing}
Ali Farhadi, Ian Endres, Derek Hoiem, and David Forsyth.
\newblock {Describing objects by their attributes}.
\newblock In {\em IEEE International Conference on Computer Vision and Pattern
  Recognition (CVPR)}, 2009.

\bibitem{Frome:2013ux}
Andrea Frome, Gregory~S Corrado, Jonathon Shlens, Samy Bengio, Jeffrey Dean,
  Marc'Aurelio Ranzato, and Tomas Mikolov.
\newblock {DeViSE: A Deep Visual-Semantic Embedding Model.}
\newblock In {\em Conference on Neural Information Processing Systems (NIPS)},
  2013.

\bibitem{Goodfellow:2014td}
I~Goodfellow, J~Pouget-Abadie, and M~Mirza.
\newblock {Generative adversarial nets}.
\newblock In {\em NIPS}, 2014.

\bibitem{gretton2012kernel}
Arthur Gretton, Karsten~M Borgwardt, Malte~J Rasch, Bernhard Sch{\"o}lkopf, and
  Alexander Smola.
\newblock A kernel two-sample test.
\newblock {\em Journal of Machine Learning Research}, 13(Mar):723--773, 2012.

\bibitem{Guo:2004dh}
Hongyu Guo and Herna~L Viktor.
\newblock {Learning from imbalanced data sets with boosting and data generation
  - the DataBoost-IM approach.}
\newblock {\em SIGKDD Explorations}, 2004.

\bibitem{jaderberg2015reading}
Max Jaderberg, Karen Simonyan, Andrea Vedaldi, and Andrew Zisserman.
\newblock {Reading Text in the Wild with Convolutional Neural Networks}.
\newblock {\em International Journal of Computer Vision}, 116(1):1--20, 2016.

\bibitem{Jayaraman:2014wq}
Dinesh Jayaraman and Kristen Grauman.
\newblock {Zero-shot recognition with unreliable attributes}.
\newblock In {\em Conference on Neural Information Processing Systems (NIPS)},
  2014.

\bibitem{kingma2014adam}
Diederik Kingma and Jimmy Ba.
\newblock Adam: A method for stochastic optimization.
\newblock {\em arXiv preprint arXiv:1412.6980}, 2014.

\bibitem{Lampert:2014fs}
Christoph~H Lampert, Hannes Nickisch, and Stefan Harmeling.
\newblock {Attribute-Based Classification for Zero-Shot Visual Object
  Categorization.}
\newblock {\em IEEE Trans Pattern Anal Mach Intell}, 36(3):453--465, 2014.

\bibitem{LeCun:1998hy}
Y~LeCun, L~Bottou, Y~Bengio, and P~Haffner.
\newblock {Gradient-based learning applied to document recognition}.
\newblock {\em Proceedings of the IEEE}, 86(11):2278--2324, November 1998.

\bibitem{li2015generative}
Yujia Li, Kevin Swersky, and Richard~S Zemel.
\newblock Generative moment matching networks.
\newblock In {\em ICML}, pages 1718--1727, 2015.

\bibitem{maas2013rectifier}
Andrew~L Maas, Awni~Y Hannun, and Andrew~Y Ng.
\newblock Rectifier nonlinearities improve neural network acoustic models.
\newblock In {\em in ICML Workshop on Deep Learning for Audio, Speech and
  Language Processing}. Citeseer, 2013.

\bibitem{makhzani2015adversarial}
Alireza Makhzani, Jonathon Shlens, Navdeep Jaitly, Ian Goodfellow, and Brendan
  Frey.
\newblock Adversarial autoencoders.
\newblock {\em arXiv preprint arXiv:1511.05644}, 2015.

\bibitem{mikolov2013distributed}
Tomas Mikolov, Ilya Sutskever, Kai Chen, Greg~S Corrado, and Jeff Dean.
\newblock Distributed representations of words and phrases and their
  compositionality.
\newblock In {\em Advances in neural information processing systems}, pages
  3111--3119, 2013.

\bibitem{Mirza:2014wi}
Mehdi Mirza and Simon Osindero.
\newblock {Conditional Generative Adversarial Nets}.
\newblock {\em arXiv}, November 2014.

\bibitem{norouzi2013zero}
Mohammad Norouzi, Tomas Mikolov, Samy Bengio, Yoram Singer, Jonathon Shlens,
  Andrea Frome, Greg~S Corrado, and Jeffrey Dean.
\newblock Zero-shot learning by convex combination of semantic embeddings.
\newblock {\em arXiv preprint arXiv:1312.5650}, 2013.

\bibitem{odena2016conditional}
Augustus Odena, Christopher Olah, and Jonathon Shlens.
\newblock Conditional image synthesis with auxiliary classifier gans.
\newblock {\em arXiv preprint arXiv:1610.09585}, 2016.

\bibitem{patterson2012sun}
Genevieve Patterson and James Hays.
\newblock Sun attribute database: Discovering, annotating, and recognizing
  scene attributes.
\newblock In {\em Computer Vision and Pattern Recognition (CVPR), 2012 IEEE
  Conference on}, pages 2751--2758. IEEE, 2012.

\bibitem{romera2015embarrassingly}
Bernardino Romera-Paredes and Philip~HS Torr.
\newblock {An embarrassingly simple approach to zero-shot learning}.
\newblock In {\em ICML}, pages 2152--2161, 2015.

\bibitem{russakovsky2015imagenet}
Olga Russakovsky, Jia Deng, Hao Su, Jonathan Krause, Sanjeev Satheesh, Sean Ma,
  Zhiheng Huang, Andrej Karpathy, Aditya Khosla, Michael Bernstein, et~al.
\newblock Imagenet large scale visual recognition challenge.
\newblock {\em International Journal of Computer Vision}, 115(3):211--252,
  2015.

\bibitem{simonyan2014very}
Karen Simonyan and Andrew Zisserman.
\newblock {Very Deep Convolutional Networks for Large-Scale Image Recognition}.
\newblock In {\em ICLR}, 2014.

\bibitem{srivastava2014dropout}
Nitish Srivastava, Geoffrey~E Hinton, Alex Krizhevsky, Ilya Sutskever, and
  Ruslan Salakhutdinov.
\newblock Dropout: a simple way to prevent neural networks from overfitting.
\newblock {\em Journal of Machine Learning Research}, 15(1):1929--1958, 2014.

\bibitem{szegedy2014going}
Christian Szegedy, Wei Liu, Yangqing Jia, Pierre Sermanet, Scott Reed, Dragomir
  Anguelov, Dumitru Erhan, Vincent Vanhoucke, and Andrew Rabinovich.
\newblock {Going deeper with convolutions}.
\newblock In {\em IEEE International Conference on Computer Vision and Pattern
  Recognition (CVPR)}, 2015.

\bibitem{Ulusoy:2005bl}
Ilkay Ulusoy and Christopher~M Bishop.
\newblock {Generative versus Discriminative Methods for Object Recognition.}
\newblock In {\em CVPR}, 2005.

\bibitem{wah2011caltech}
Catherine Wah, Steve Branson, Peter Welinder, Pietro Perona, and Serge
  Belongie.
\newblock The caltech-ucsd birds-200-2011 dataset.
\newblock 2011.

\bibitem{wang2016zero}
Qian Wang and Ke~Chen.
\newblock Zero-shot visual recognition via bidirectional latent embedding.
\newblock {\em arXiv preprint arXiv:1607.02104}, 2016.

\bibitem{weston2011wsabie}
Jason Weston, Samy Bengio, and Nicolas Usunier.
\newblock Wsabie: scaling up to large vocabulary image annotation.
\newblock In {\em Proceedings of the Twenty-Second international joint
  conference on Artificial Intelligence-Volume Volume Three}, pages 2764--2770.
  AAAI Press, 2011.

\bibitem{xian2016latent}
Yongqin Xian, Zeynep Akata, Gaurav Sharma, Quynh Nguyen, Matthias Hein, and
  Bernt Schiele.
\newblock Latent embeddings for zero-shot classification.
\newblock In {\em Proceedings of the IEEE Conference on Computer Vision and
  Pattern Recognition}, pages 69--77, 2016.

\bibitem{xian2017zero}
Yongqin Xian, Bernt Schiele, and Zeynep Akata.
\newblock Zero-shot learning-the good, the bad and the ugly.
\newblock In {\em IEEE Conference on Computer Vision and Pattern Recognition},
  2017.

\bibitem{Zhang:2015vs}
Ziming Zhang and Venkatesh Saligrama.
\newblock {Zero-Shot Learning via Semantic Similarity Embedding.}
\newblock In {\em IEEE International Conference on Computer Vision (ICCV)},
  2015.

\bibitem{zhang2016zero}
Ziming Zhang and Venkatesh Saligrama.
\newblock Zero-shot learning via joint latent similarity embedding.
\newblock In {\em Proceedings of the IEEE Conference on Computer Vision and
  Pattern Recognition}, pages 6034--6042, 2016.

\bibitem{zhou2014learning}
Bolei Zhou, Agata Lapedriza, Jianxiong Xiao, Antonio Torralba, and Aude Oliva.
\newblock Learning deep features for scene recognition using places database.
\newblock In {\em Advances in neural information processing systems}, pages
  487--495, 2014.

\end{thebibliography}
}
\end{document}